\documentclass[letterpaper, 10 pt, conference]{ieeeconf}

\IEEEoverridecommandlockouts
\usepackage{amsmath,amssymb,amsfonts}
\usepackage[ruled]{algorithm}
\usepackage{algorithmic}
\usepackage{graphicx}
\usepackage{flafter}
\usepackage{textcomp}
\usepackage{url}
\usepackage{booktabs}
\usepackage{multirow}
\usepackage{tabularx}
\usepackage[table]{xcolor}
\usepackage{placeins}
\usepackage{flushend}
\usepackage{capt-of}
\usepackage{etoolbox}
\usepackage[protrusion=true,expansion=true]{microtype}
\AtBeginEnvironment{thebibliography}{\flushbottom}
\AtBeginDocument{\addtolength{\topskip}{3pt}}
\apptocmd{\normalsize}{\setlength{\baselineskip}{12pt plus 0.3pt}}{}{}
\AtBeginDocument{%
  \setlength{\abovedisplayskip}{1.5ex plus 1pt}%
  \setlength{\belowdisplayskip}{1.5ex plus 1pt}%
  \setlength{\abovedisplayshortskip}{0pt}%
  \setlength{\belowdisplayshortskip}{1.5ex plus 1pt}}

\makeatletter
\renewcommand{\section}{\@startsection{section}{1}{\z@}%
{1.5ex}{0.7ex}{\normalfont\normalsize\centering\scshape}}
\renewcommand{\paragraph}{\@startsection{paragraph}{4}{2\parindent}%
{0ex}{0ex}{\normalfont\normalsize\bfseries\itshape}}
\renewcommand{\subsection}{\@startsection{subsection}{2}{\z@}%
{1.5ex}{0.7ex}{\normalfont\normalsize\itshape}}
\newcommand{\cmapkeepheading}{\par\begingroup
  \@tempdima=\pagegoal\advance\@tempdima by -\pagetotal
  \ifdim\@tempdima<4\baselineskip\penalty-10000\relax\fi\endgroup}
\makeatother

\makeatletter
\providecommand{\bstctlcite}{\@ifnextchar[{\@bstctlcite}{\@bstctlcite[@auxout]}}
\def\@bstctlcite[#1]#2{\@bsphack
  \@for\@citeb:=#2\do{%
    \edef\@citeb{\expandafter\@firstofone\@citeb}%
    \if@filesw\immediate\write\csname #1\endcsname{\string\citation{\@citeb}}\fi}%
  \@esphack}
\makeatother

\makeatletter
\let\cmap@originalmakecaption\@makecaption
\long\def\@makecaption#1#2{%
  \ifx\@captype\@IEEEtablestring
    \begingroup
      \centering\footnotesize
      #1\par\nobreak\vskip2pt
      {\scshape #2\par}%
    \endgroup
    \nobreak\vskip3pt
  \else
    \cmap@originalmakecaption{#1}{#2}%
  \fi}
\makeatother

\newcommand{\R}{\mathbb{R}}

\newcommand{\rev}[1]{\textcolor{black}{#1}}
\newenvironment{revision}{\begingroup\color{black}}{\endgroup}
\definecolor{cmapaccent}{RGB}{44,92,123}
\definecolor{cmaprowbg}{RGB}{232,244,247}
\let\citep\cite
\let\citet\cite

\newcolumntype{Y}{>{\centering\arraybackslash}X}
\newcolumntype{L}[1]{>{\raggedright\arraybackslash}p{#1}}
\newcolumntype{C}[1]{>{\centering\arraybackslash}p{#1}}

\newcommand{\fullerpar}{\par}

\newcommand{\insertmethodfigure}{%
\begin{figure*}[!t]
\centering
\vspace*{3pt}
\includegraphics[width=\textwidth]{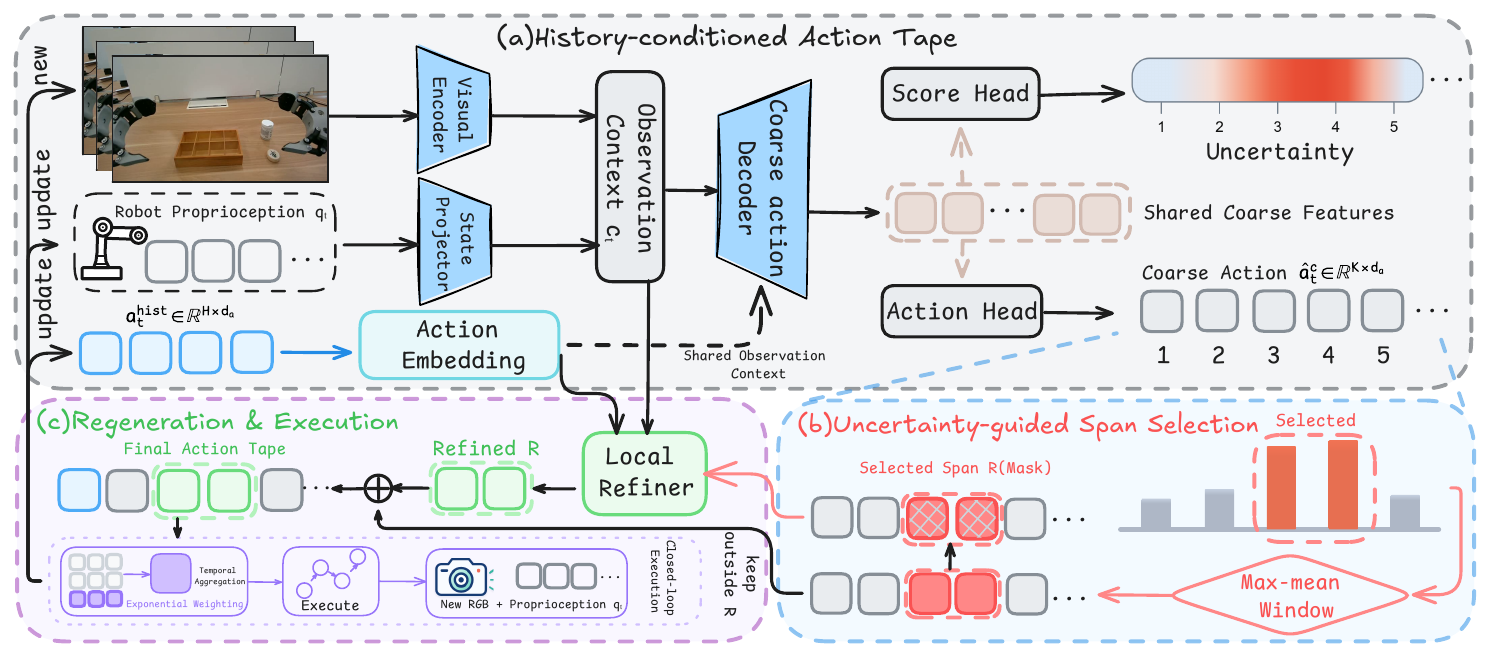}
\vspace{-21pt}
\caption{\rev{Overview of CMAP. (a) HCAT combines executed history and masked future positions on one action tape, using visual and proprioceptive context to predict coarse actions and error-supervised uncertainty scores. UGSR comprises (b) uncertainty-guided selection of a contiguous span $\mathcal{R}$ and (c) regeneration and replacement of that span, with all surrounding actions copied unchanged. Only issued commands update the execution history.}}
\label{fig:method}
\end{figure*}}

\newcommand{\insertsensitivityfigure}{%
\begin{figure}[t]
\cmapsensitivityfloat
\centering
\vspace{-5pt}
\includegraphics[width=0.91\columnwidth]{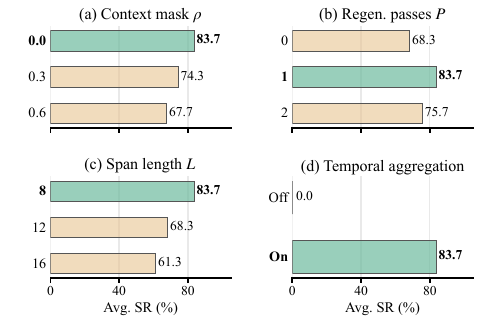}
\vspace{-12pt}
\caption{\rev{CMAP sensitivity on three RoboTwin tasks. Bars report macro-average success. Green marks default settings.}}
\label{fig:cmap_sensitivity}
\end{figure}}

\newcommand{\insertlocalregenerationfigure}{%
\begin{figure}[t]
\cmaplocalfloat
\centering
\vspace*{10pt}
\includegraphics[width=\columnwidth]{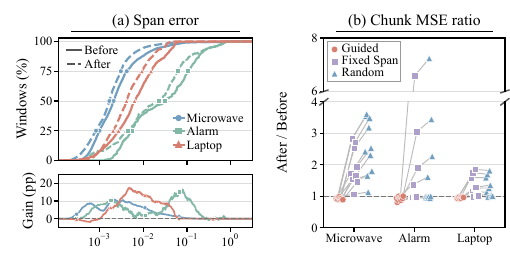}\par
\vspace{-6pt}
\caption{\rev{Local regeneration analysis. (a) Cumulative span-error distributions before and after regeneration, with After-minus-Before differences below. (b) Paired chunk-error ratios across selection strategies. Ratios below one indicate improvement.}}
\label{fig:local_regeneration}
\end{figure}}

\makeatletter
\let\cmap@addtonextcol\@addtonextcol
\newcommand{\cmapfloatat}[2]{%
  \expandafter\xdef\csname cmap@page@\number\@currbox\endcsname{#1}%
  \expandafter\xdef\csname cmap@col@\number\@currbox\endcsname{#2}}
\newcommand{\cmaprightcolumnfloat}{\cmapfloatat{7}{0}}
\newcommand{\cmapsensitivityfloat}{\cmapfloatat{6}{0}}
\newcommand{\cmaprevisionrulefloat}{\cmapfloatat{6}{0}}
\newcommand{\cmapsimulationfloat}{\cmapfloatat{4}{1}}
\newcommand{\cmaplocalfloat}{\cmapfloatat{6}{1}}
\newcommand{\cmapalgorithmfloat}{\cmapfloatat{4}{0}}
\apptocmd{\fs@ruled}{\pretocmd{\@fs@pre}{\vskip10pt}{}{}}{}{}
\newcommand{\cmaptabletwofloat}{\cmapfloatat{5}{0}}
\newcommand{\cmaptablethreefloat}{\cmapfloatat{5}{1}}
\def\@addtonextcol{%
  \ifcsname cmap@page@\number\@currbox\endcsname
    \ifnum\c@page<\csname cmap@page@\number\@currbox\endcsname\relax
      \@cons\@deferlist\@currbox
    \else
      \ifnum\csname cmap@col@\number\@currbox\endcsname=\if@firstcolumn0\else1\fi\relax
        \cmap@addtonextcol
        \if@insert
          \expandafter\global\expandafter\let
            \csname cmap@page@\number\@currbox\endcsname\@undefined
        \fi
      \else
        \@cons\@deferlist\@currbox
      \fi
    \fi
  \else
    \cmap@addtonextcol
  \fi}
\makeatother

\newcommand{\cmaprobotgroupline}[2]{%
  \hbox to #1{\kern2pt%
    \vrule width0.35pt height4.7pt depth0.1pt%
    \leaders\hrule height2.6pt depth-2.25pt\hfill%
    \kern4pt{\fontsize{8}{9}\selectfont\bfseries #2}\kern4pt%
    \leaders\hrule height2.6pt depth-2.25pt\hfill%
    \vrule width0.35pt height4.7pt depth0.1pt\kern2pt}}
\newcommand{\cmaprealworldgraphic}{%
  \begin{minipage}{0.99\textwidth}
  \offinterlineskip
  \hbox{\includegraphics[width=\linewidth,trim=0 142.2bp 0 0,clip]{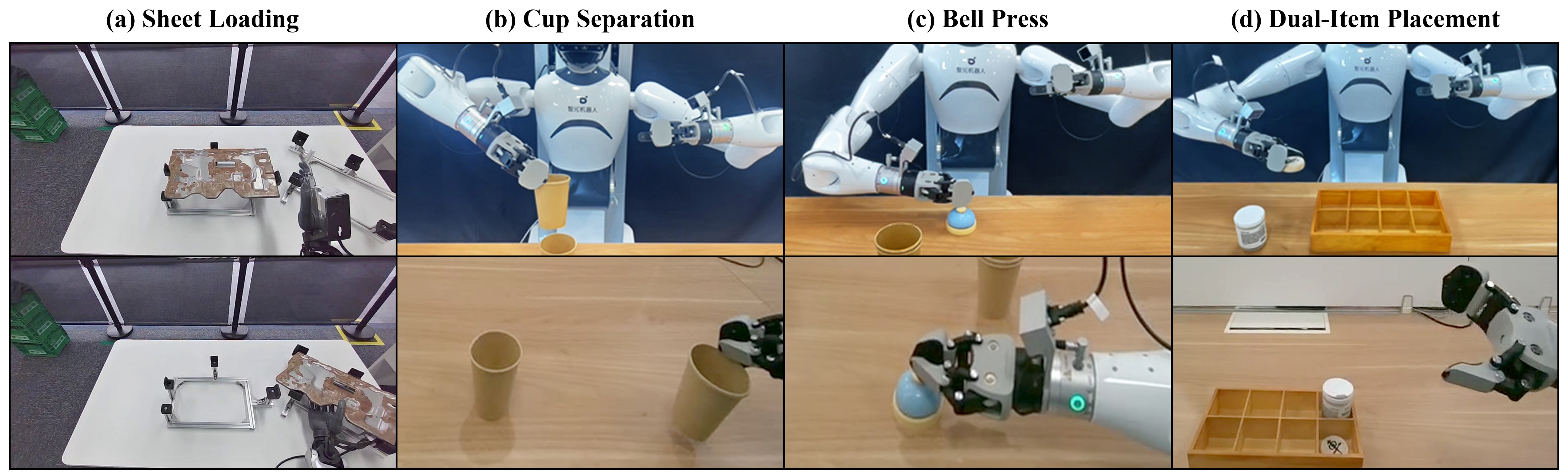}}%
  \vskip1pt
  \hbox to\linewidth{\hskip0.005586\linewidth%
    \cmaprobotgroupline{0.247207\linewidth}{Leju Kuavo 4 Pro}%
    \cmaprobotgroupline{0.741620\linewidth}{AGIBOT G1}\hss}%
  \vskip3pt
  \hbox{\includegraphics[width=\linewidth,trim=0 0 0 13.92bp,clip]{figures/paper/fig_realworld_rollouts_double_column_row_clear_v4.png}}%
  \end{minipage}}

\newcommand{\insertrealworldrollouts}{%
\begin{figure*}[t]
\centering
\vspace*{10pt}
\setlength{\abovecaptionskip}{2pt}
\cmaprealworldgraphic\par
\caption{\rev{Representative physical rollouts on Kuavo 4 Pro and AGIBOT G1. Each column pairs the task context with its final outcome.}}
\label{fig:realworld_rollouts}
\end{figure*}}

\newcommand{\insertrealworldtable}{%
\begin{table}[!t]
\cmaprightcolumnfloat
\centering
\caption{\rev{Real-world success counts (10 trials per task). Avg. is the mean of five success rates.}}
\label{tab:realworld_main}
\scriptsize
\renewcommand{\arraystretch}{1.08}
\setlength{\tabcolsep}{1.4pt}
\begin{tabularx}{\linewidth}{@{}lYYYYYY@{}}
\toprule
\multirow{3}{*}{\textbf{Method}}
& \multicolumn{4}{c}{\scriptsize\bfseries AGIBOT}
& \multicolumn{1}{c}{\scriptsize\bfseries Kuavo}
& \multirow{3}{*}{\shortstack{\textbf{Avg.}\\\textbf{(\%)}}} \\
\cmidrule(lr){2-5}\cmidrule(lr){6-6}
& \multirow{2}{*}{\shortstack{Cup\\Sep.}}
& \multirow{2}{*}{\shortstack{Bell\\Press}}
& \multicolumn{2}{c}{\textbf{Dual-item Placement}}
& \multirow{2}{*}{\shortstack{Sheet\\Load.}}
& \\
\cmidrule(lr){4-5}
& & & \shortstack{Chess\\Grasp} & \shortstack{Bottle\\Grasp} & & \\
\midrule
ACT & 4/10 & 8/10 & 5/10 & 6/10 & 2/10 & 50.0 \\
DP & 5/10 & 7/10 & 3/10 & 8/10 & 1/10 & 48.0 \\
DDP & 4/10 & 6/10 & 5/10 & 5/10 & 2/10 & 44.0 \\
\rowcolor{cmaprowbg}
\textbf{CMAP} & \textbf{7/10} & \textbf{9/10} & \textbf{8/10} & \textbf{10/10} & \textbf{4/10} & \textbf{76.0} \\
\bottomrule
\end{tabularx}
\end{table}}

\newcommand{\insertcomparisonspage}{%
\begin{table*}[!t]
\centering
\vspace*{10pt}
\begin{minipage}{\textwidth}
\centering
\captionof{table}{\rev{RoboTwin success rates (\%) on six tasks. RDT-1B and TwinVLA results are taken from their original papers. Column maxima are bolded, with ties highlighted only for CMAP. The CMAP row is shaded.}}
\label{tab:robotwin_main}
\small
\renewcommand{\arraystretch}{1.04}
\setlength{\tabcolsep}{3.2pt}
\begin{revision}
\begin{tabularx}{\linewidth}{@{}L{0.18\linewidth}YYYYYY|C{0.055\linewidth}@{}}
\toprule
\textbf{Method}
& \textbf{Handover}
& \textbf{Stack Blocks}
& \textbf{Microwave}
& \textbf{Alarm Clock}
& \textbf{Laptop}
& \textbf{Empty Cup}
& \textbf{Avg.} \\
\midrule
ACT~\citep{zhao2023act} & 83.0 & 19.0 & 65.0 & 29.0 & 60.0 & 45.0 & 50.2 \\
Diffusion Policy~\citep{chi2025diffusion} & 53.0 & 7.0 & 5.0 & 61.0 & 49.0 & 37.0 & 35.3 \\
DDP~\citep{mgp} & 93.0 & 10.0 & 30.0 & 18.0 & 57.0 & 14.0 & 37.0 \\
RDT-1B~\citep{liu2024rdt} & 90.0 & 21.0 & 37.0 & 61.0 & 59.0 & 56.0 & 54.0 \\
TwinVLA~\citep{im2026twinvla} & 84.0 & 26.0 & 3.0 & 33.0 & 80.0 & 50.0 & 46.0 \\
\rowcolor{cmaprowbg}
\textbf{CMAP (Ours)} & \textbf{97.0} & \textbf{26.0} & \textbf{92.0} & \textbf{74.0} & \textbf{85.0} & \textbf{56.0} & \textbf{71.7} \\
\bottomrule
\end{tabularx}
\end{revision}
\end{minipage}
\end{table*}}

\newcommand{\insertcomparisoncolumns}{%
\begin{table}[!t]
\cmaptabletwofloat
\centering
\caption{\rev{ManiSkill success rates (\%). Bold marks CMAP's best or tied-best results. $\dagger$ marks published results from StaMo~\citep{liu2026stamo}.}}
\label{tab:maniskill_comparison}
\scriptsize
\renewcommand{\arraystretch}{1.04}
\setlength{\tabcolsep}{3.2pt}
\begin{tabularx}{\linewidth}{@{}L{0.31\linewidth}YYY|C{0.08\linewidth}@{}}
\toprule
\textbf{Method} & \mbox{Push Cube} & \mbox{Stack Cube} & \mbox{Pull Cube} & \textbf{Avg.} \\
\midrule
ACT~\citep{zhao2023act} & 31.0 & 13.0 & 34.0 & 26.0 \\
OpenVLA~\citep{kim2024openvla}$^\dagger$ & 4.0 & 18.0 & 16.0 & 12.7 \\
StaMo~\citep{liu2026stamo}$^\dagger$ & 32.0 & 16.0 & 38.0 & 28.7 \\
\rowcolor{cmaprowbg}
\textbf{CMAP (Ours)} & \textbf{54.0} & \textbf{18.0} & \textbf{88.0} & \textbf{53.3} \\
\bottomrule
\end{tabularx}
\vspace{7pt}
\end{table}
}

\newcommand{\insertcomponenttablepage}{%
\begin{table}[t]
\cmaptablethreefloat
\centering
\caption{\rev{Component ablation success rates (\%). +Regen uses random spans.}}
\label{tab:component_buildup}
\scriptsize
\renewcommand{\arraystretch}{1.08}
\setlength{\tabcolsep}{2.7pt}
\begin{revision}
\begin{tabularx}{\linewidth}{lYYYYY}
\toprule
\textbf{Variant} & \textbf{HCAT} & \textbf{Span regen.} & \textbf{UGSR}
& \textbf{RoboTwin} & \textbf{ManiSkill} \\
\midrule
ACT & -- & -- & -- & 51.3 & 26.0 \\
+History & \checkmark & -- & -- & 65.0 & 45.3 \\
+Regen & \checkmark & \checkmark & -- & 66.7 & 43.0 \\
\rowcolor{cmaprowbg}
\textbf{CMAP} & \checkmark & \checkmark & \checkmark & \textbf{83.7} & \textbf{53.3} \\
\bottomrule
\end{tabularx}
\end{revision}
\vspace{8.34pt}
\end{table}}

\newcommand{\insertpagefourfigures}{%
\begin{figure}[t]
\cmapsimulationfloat
\centering
\vspace*{10pt}
\includegraphics[width=\linewidth]{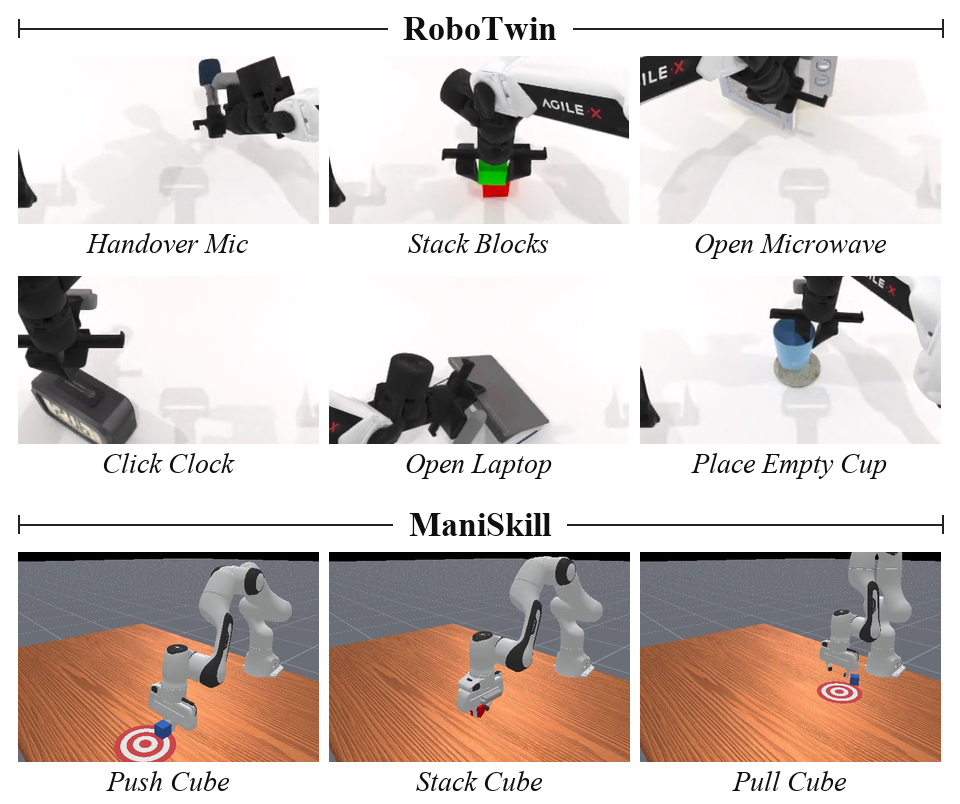}
\vspace{-18pt}
\caption{\rev{Simulation tasks with six RoboTwin tasks at the top and three ManiSkill tasks at the bottom.}}
\label{fig:simulation_tasks}
\end{figure}
}

\newcommand{\insertrevisionrulefigure}{%
\begin{figure}[!t]
\cmaprevisionrulefloat
\centering
\vspace*{10pt}
\setlength{\abovecaptionskip}{0pt}
\includegraphics[width=\linewidth,trim=0 6bp 0 0,clip]{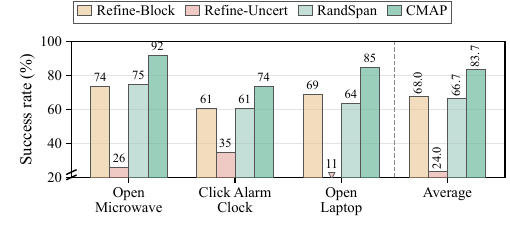}\par
\vspace{1pt}
\caption{\rev{RoboTwin revision-rule ablation. Axis starts at 20\%. The downward marker denotes 11\%. The dashed line separates the macro-average.}}
\label{fig:ablation}
\end{figure}
}

\def\BibTeX{{\rm B\kern-.05em{\sc i\kern-.025em b}\kern-.08em
    T\kern-.1667em\lower.7ex\hbox{E}\kern-.125emX}}
\begin{document}
\bstctlcite{cmap_bst_control}

\title{\LARGE \bf
Predict in Chunks, Revise in Spans:\\
History-conditioned Action Regeneration for Robotic Manipulation
}

\author{Chenyang Wang$^{1,*}$, Yuntian Wang$^{1}$, Jiaying Tang$^{1,\dagger}$,
Chunzheng Li$^{1,\dagger}$, Yang Liu$^{2}$, and Siao Liu$^{1}$\\[2pt]
{\small $^{1}$Soochow University, China \qquad $^{2}$Tongji University, China}\\[1pt]
{\small $^{\dagger}$Equal contribution \qquad $^{*}$Corresponding author}%
\thanks{This work has been submitted to the IEEE for possible publication. Copyright may be transferred without notice, after which this version may no longer be accessible.}
}

\maketitle
\thispagestyle{empty}
\pagestyle{empty}

\begin{abstract}
Action chunking improves temporal coherence in visuomotor imitation learning by jointly predicting sequences of future actions. However, many existing policies regenerate entire action chunks even when errors are localized, potentially replacing otherwise valid predictions. We introduce Contextual Masked Action Policy (CMAP), a visuomotor policy that refines continuous action chunks through history-conditioned local regeneration. Specifically, CMAP introduces a history-conditioned action tape that combines executed commands with masked future positions to produce coarse action predictions and per-step uncertainty scores. An uncertainty-guided span regeneration mechanism then selects a high-uncertainty contiguous span and reconstructs it using the executed history and retained future predictions as context. By jointly reconstructing neighboring actions, CMAP models local temporal dependencies while preserving all predictions outside the selected span. Extensive experiments on nine simulated and four real-world manipulation tasks demonstrate that CMAP achieves superior overall performance over several competitive baselines. Ablation studies validate the complementary contributions of its key components, while held-out trajectory analysis shows that local regeneration reduces prediction errors within selected spans.
\fullerpar

\end{abstract}


\section{Introduction}
\label{sec:intro}
Learning visuomotor policies from demonstrations requires mapping high-dimensional observations to temporally coordinated action sequences. During closed-loop execution, even small prediction errors can drive the policy toward states poorly covered by the demonstrations, causing errors to accumulate over time~\citep{ross2011reduction}. Action chunking mitigates this issue by jointly predicting short sequences of future actions, improving temporal coherence over single-step prediction~\citep{zhao2023act,chi2025diffusion}. It has consequently become a mainstream approach in visuomotor imitation learning. However, this temporal coherence comes at the cost of coarse-grained correction. Because prediction and revision operate at the chunk level, an error confined to only part of the sequence can force the entire chunk to be reconsidered. Existing methods therefore often regenerate the whole chunk even when most predicted actions remain usable, extending the revision beyond the region that requires correction within the predicted chunk.

\begin{figure}[!t]
\centering
\includegraphics[width=\linewidth]{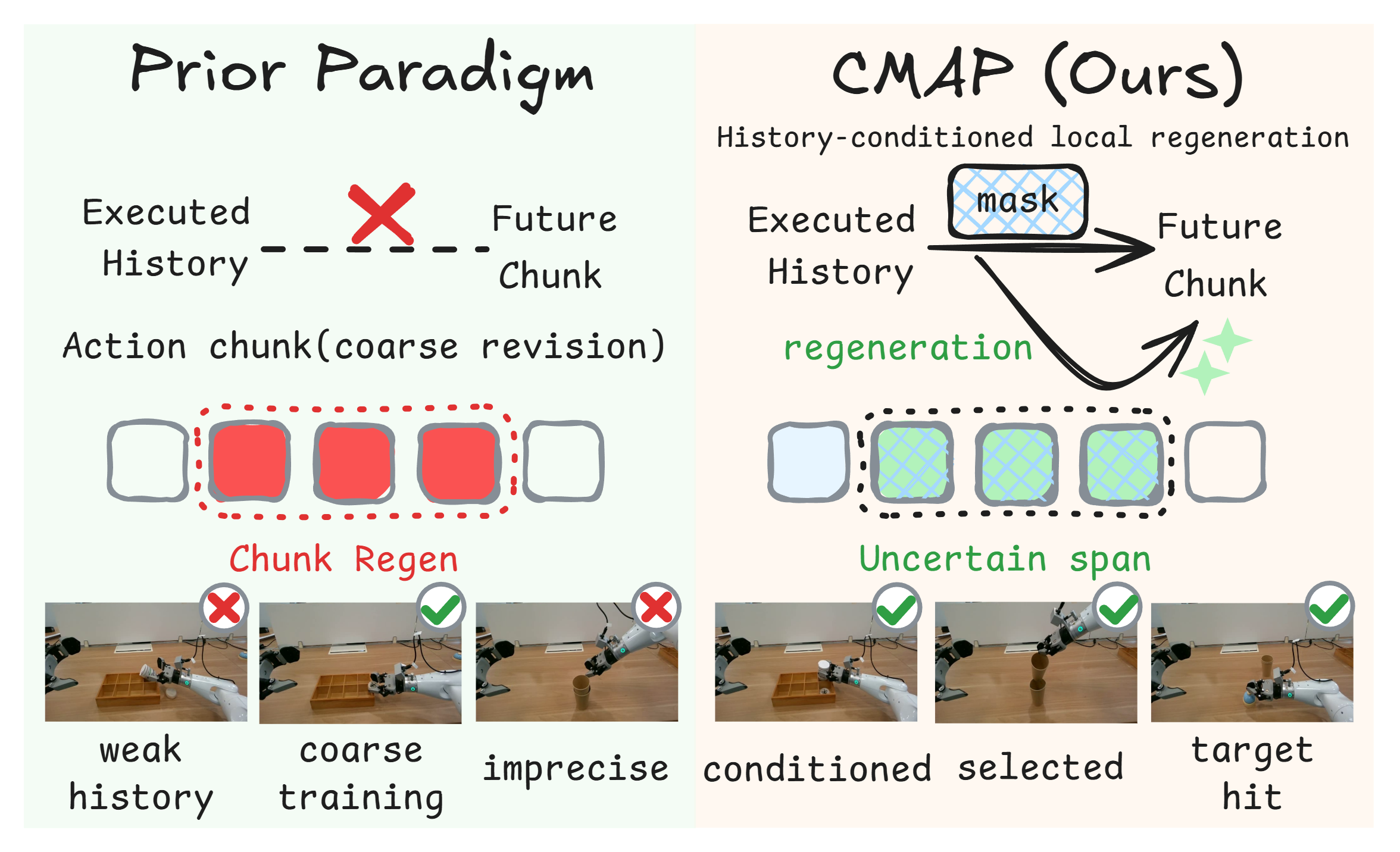}
\vspace{-21pt}
\caption{\rev{Whole-chunk prediction versus CMAP's editable action tape. Conventional policies regenerate or accept the complete chunk as one object. CMAP uses executed history to localize an unreliable contiguous span, regenerate only that span, and preserve reliable surrounding actions.}}
\label{fig:concept}
\end{figure}

Recent advances in masked generation provide a promising direction toward localized revision of action chunks. In discrete action spaces, Masked Generative Policy (MGP)~\citep{mgp} iteratively refines low-confidence action tokens, while LLaDA-VLA enables localized action-token prediction through structured decoding in a diffusion VLM~\citep{lladavla}. These approaches demonstrate the potential of selective revision, but rely on discrete action representations that do not naturally match continuous robot control. For continuous control, diffusion-based methods have explored trajectory inpainting and masked refinement as mechanisms for partially constraining or revising generated action sequences. For instance, Hierarchical Diffusion Policy (HDP) uses trajectory inpainting to generate continuous motions under prescribed constraints~\citep{hdp}, while ReMAC introduces masked action chunking to correct locally inconsistent actions during execution~\citep{remac}. However, existing methods typically rely on predefined or heuristic correction regions, which may leave unreliable actions uncorrected or cause unnecessary changes to otherwise usable predictions.

We attribute this limitation to two key factors. First, the reliability of a predicted action depends on its execution context, yet most chunked policies primarily condition on the current observation without explicitly retaining the actions already executed by the robot. Consequently, visually similar observations reached through different action histories require different continuations, making it difficult to assess whether a newly predicted chunk is consistent with the realized motion. Second, continuous actions are temporally coupled, making independent timestep-wise correction insufficient for preserving motion consistency. Pointwise refinement may fragment a coherent action sequence and fail to capture an unreliable region that spans multiple neighboring timesteps. Effective localized revision therefore requires history-aware estimation of action reliability and span-level correction that retains all future predictions outside the selected interval.

To address these issues, we propose \textbf{C}ontextual \textbf{M}asked \textbf{A}ction \textbf{P}olicy (\textbf{CMAP}), a history-conditioned framework that replaces whole-chunk regeneration with localized editing of continuous action chunks, as illustrated in Fig.~\ref{fig:concept}. CMAP introduces a unified \emph{action tape} that combines recently executed commands with masked future action positions. First, the \textbf{H}istory-\textbf{c}onditioned \textbf{A}ction \textbf{T}ape (\textbf{HCAT}) incorporates executed-action history into coarse action prediction and per-step uncertainty estimation. Second, the \textbf{U}ncertainty-\textbf{g}uided \textbf{S}pan \textbf{R}egeneration (\textbf{UGSR}) aggregates uncertainty over contiguous candidate intervals, identifies unreliable spans, and selectively regenerates only the affected region while preserving the remaining predictions. We evaluate CMAP on RoboTwin~\cite{mu2024robotwin}, ManiSkill~\citep{gu2023maniskill2}, and two real-world platforms, AGIBOT G1 and Leju Kuavo 4 Pro. CMAP consistently achieves the best overall performance against representative action-chunking and diffusion-based baselines across these settings. Controlled ablations further validate the complementary benefits of executed-action history and span-level regeneration, particularly for tasks involving localized execution errors. Our contributions address continuous action representation, executed-action context, and uncertainty-guided revision.
\begin{itemize}
\item \textbf{Editable continuous action chunks.} We formulate chunk prediction as localized revision that preserves reliable continuous actions outside the selected interval.

\item \textbf{History-conditioned Action Tape.} HCAT anchors the next chunk to commands already issued by the robot and jointly produces coarse actions and uncertainty scores.

\item \textbf{Uncertainty-guided Span Regeneration.} UGSR addresses localized action errors by regenerating a contiguous span selected using per-step uncertainty scores. Joint reconstruction supports temporally coherent revision while preserving all predictions outside the span.\fullerpar
\end{itemize}

\section{Related Work}
\label{sec:related}

\subsection{\rev{Chunking-based Visuomotor Policies}}
\begin{revision}
ACT~\citep{zhao2023act} predicts action chunks with a conditional variational autoencoder and combines overlapping predictions through temporal aggregation. Extensions support mobile manipulation~\citep{fu2024mobile} and tactile-conditioned bimanual control~\citep{gu2025tactilealoha}. Diffusion Policy~\citep{chi2025diffusion} generates continuous trajectories through iterative denoising under receding-horizon control. Movement Primitive Diffusion~\citep{scheikl2024mpd} combines diffusion with probabilistic dynamic movement primitives to generate smooth trajectories. VQ-BeT~\citep{lee2024vqbet} represents actions through a learned discrete bottleneck, while autoregressive approaches use chunking causal transformers~\citep{zhang2025autoregressive}. Larger policies, including $\pi_0$~\citep{black2024pi0}, RDT-1B~\citep{liu2024rdt}, and ManiFlow~\citep{yan2025maniflow}, further explore diffusion or flow-based action generation. These approaches model dependencies across future actions and update motion through aggregation or replanning.
\end{revision}

\subsection{\rev{Masked Action Modeling}}
\begin{revision}
Beyond directly predicting complete sequences, masked generation supports conditional completion and iterative refinement~\citep{ghazvininejad2019maskpredict,chang2022maskgit,carroll2022unimask,wu2023mtm}. MGP~\citep{mgp} represents actions as discrete tokens and dynamically remasks low-confidence predictions, whereas LLaDA-VLA~\citep{lladavla} adapts a diffusion VLM through localized action-token classification and hierarchical action-structured decoding. Discrete Diffusion VLA~\citep{discretevla} and TBD-VLA~\citep{tbdvla} further study discrete diffusion or block-wise action decoding in VLA models. In continuous action space, RTC~\citep{rtc} and ReMaC~\citep{remac} perform action inpainting under partial action conditioning, preserving committed actions while completing editable future positions. ARP~\citep{wang2026arp} couples visual--action alignment with two-step continuous residual refinement to improve decoded-skill execution precision. Heteroscedastic objectives, ensembles, and dropout-based methods provide uncertainty estimates~\citep{kendall2017uncertainty,lakshminarayanan2017simple,gal2016dropout} that support intervention~\citep{menda2019ensembledagger,hoque2022thriftydagger}, failure detection~\citep{xu2025failure}, and exploration in residual reinforcement learning~\citep{dodeja2026residual}. UGR~\citep{wang_ugr} applies uncertainty-guided sparse residual correction at selected positions within predicted action chunks, allowing updates at temporally separated action positions.\fullerpar
\end{revision}

\subsection{\rev{History-conditioned Policy Learning}}
\begin{revision}
\insertmethodfigure

Visuomotor policies encode temporal context through observation histories~\citep{mandlekar2022matters,chi2025diffusion} or recurrent latent states~\citep{zhou2025mtil}. During asynchronous execution, RTC~\citep{rtc} and ReMaC~\citep{remac} condition continuous action inpainting on a committed prefix to connect a new prediction with commands scheduled for execution. Observation history captures changes in the scene, whereas action-prefix conditioning specifies the motion that a new prediction must continue. This distinction is relevant when similar observations follow different command sequences. CMAP combines executed-action history with uncertainty-guided local regeneration in continuous action space. It selects and reconstructs a contiguous future span while retaining the remaining predictions as context for masked reconstruction.\fullerpar
\end{revision}

\section{Method}
\label{sec:method}

\subsection{\rev{Overview}}
\label{sec:problem_setup}
\begin{revision}
Given the current visual and proprioceptive observations and a sequence of previously issued commands, CMAP predicts a future action chunk and selectively regenerates a contiguous part of it. Fig.~\ref{fig:method} illustrates the computation from observation encoding to action execution. Executed commands and future action positions are arranged in a single temporal sequence, which we refer to as an \emph{action tape}. Each position represents one continuous action vector. The historical positions provide the action context, while the future positions are initially masked and subsequently populated with predictions. A transformer produces coarse actions together with uncertainty scores, which determine the interval to be masked and regenerated. The regenerated actions replace the selected interval, and the resulting chunk supplies the commands for subsequent closed-loop execution.\fullerpar

We learn from offline demonstrations $\mathcal D=\{\tau_n\}_{n=1}^{M}$. At timestep $t$, let $I_t$ denote the visual observations, $q_t$ the proprioceptive state, and \mbox{$\mathbf a_t^{\mathrm{hist}}=[a_{t-H},\ldots,a_{t-1}]$} the preceding $H$ issued commands, with $a_t\in\R^{d_a}$. The policy predicts $K$ future actions,
\begin{equation}
\hat{\mathbf a}_t^{\mathrm{fut}}=\pi_\theta(I_t,q_t,\mathbf a_t^{\mathrm{hist}})\in\R^{K\times d_a}.
\label{eq:policy_definition}
\end{equation}
The demonstration sequence $\mathbf a_t^{\mathrm{fut}}=[a_t,\ldots,a_{t+K-1}]$ provides the training target. During execution, the policy is repeatedly queried with updated observations and issued-command history to generate the next future action chunk.\fullerpar
\end{revision}

\subsection{\rev{History-conditioned Action Tape}}
\label{sec:masked_tape}
\begin{revision}
As shown in Fig.~\ref{fig:method}(a), HCAT represents the historical and future actions on a shared tape,
\begin{equation}
\mathbf x_t=[\mathbf a_t^{\mathrm{hist}}\mid\mathbf a_t^{\mathrm{fut}}]
\in\R^{(H+K)\times d_a}.
\label{eq:action_tape}
\end{equation}
For coarse prediction, the historical prefix contains the issued commands and all future positions are masked. We project each visible action into the model embedding space and replace each masked action with a learned embedding. Temporal position embeddings distinguish the ordered action positions, giving
\begin{equation}
\mathbf z_i=\mathbf e_i+(1-m_i)\phi_a(x_{t,i})+m_i\mathbf e_{\mathrm{mask}},
\label{eq:tape_token}
\end{equation}
where $\phi_a$ is the action projection, $\mathbf e_i$ is the positional embedding, and $m_i$ indicates whether position $i$ is masked. This representation preserves a common temporal ordering for the executed prefix and future continuation during both coarse prediction and local span regeneration operations.\fullerpar

ResNet-18 visual encoders~\citep{he2016resnet} and a learned proprioceptive projector encode $I_t$ and $q_t$ into the observation context $c_t$. Conditioned on $c_t$, the transformer processes the tape embeddings and produces temporal features $h_i$. Two prediction heads map the future features to coarse actions and uncertainty scores,
\begin{equation}
\hat a_i^c=g_a(h_{H+i}),\quad v_i=g_s(h_{H+i}),\quad i=0,\ldots,K-1.
\label{eq:hcat_heads}
\end{equation}
\noindent Here, $g_a$ predicts continuous commands, while $g_s$ estimates uncertainty at each future position. The scores are supervised by coarse-action prediction errors and guide subsequent span selection. Both heads share temporal features conditioned on the executed history and current observation context.\fullerpar
\begin{algorithm}[t]
\cmapalgorithmfloat
\caption{\rev{CMAP training and inference}}
\label{alg:cmap}
\begin{revision}
\fontsize{9}{11.5}\selectfont
\begin{algorithmic}[1]
\REQUIRE observation context $c_t$, history $\mathbf a_t^{\mathrm{hist}}$, span length $L$, passes $P$; target $\mathbf y=\mathbf a_t^{\mathrm{fut}}$ for training
\STATE \textbf{Shared prediction and regeneration} (training: $P=1$)
\STATE $(\hat{\mathbf a}^{c},\mathbf v)\leftarrow\mathrm{HCAT}(c_t,\mathbf a_t^{\mathrm{hist}})$
\STATE $\hat{\mathbf a}\leftarrow\hat{\mathbf a}^{c}$; $\mathcal S\leftarrow\varnothing$
\FOR{$p=1$ to $P$}
\STATE $\mathcal W_j=\{j,\ldots,j+L-1\}$, $0\le j\le K-L$
\STATE $\mathcal J\leftarrow\{j:\mathcal W_j\cap\mathcal S=\varnothing\}$
\STATE \textbf{if} $\mathcal J=\varnothing$ \textbf{then break}
\STATE $j^\star\leftarrow\min\arg\max_{j\in\mathcal J}\frac1L\sum_{i\in\mathcal W_j}v_i$
\STATE $\mathcal R\leftarrow\mathcal W_{j^\star}$; $\tilde{\mathbf x}\leftarrow\operatorname{Mask}_{\mathcal R}([\mathbf a_t^{\mathrm{hist}}\mid\hat{\mathbf a}])$
\STATE $\hat{\mathbf a}^{r}_{\mathcal R}\leftarrow\mathrm{UGSR}(c_t,\tilde{\mathbf x})$
\STATE $\hat{\mathbf a}_{\mathcal R}\leftarrow\hat{\mathbf a}^{r}_{\mathcal R}$ \hfill (all other entries unchanged)
\STATE $\mathcal S\leftarrow\mathcal S\cup\mathcal R$; refresh $\mathbf v$ before the next pass
\ENDFOR
\STATE \textbf{Training:} $N$ valid future steps; $\operatorname{sg}$ stops gradients
\STATE $r_i\leftarrow\delta(y_i,\operatorname{sg}(\hat a_i^c))$; $\mathcal L_{\mathrm{score}}\leftarrow\frac1N\sum_i\ell_s(v_i,r_i)$
\STATE $\mathcal L_{\mathrm{act}}\leftarrow\ell_1(\mathbf y,\hat{\mathbf a}^{c})+\ell_1(\mathbf y_{\mathcal R},\hat{\mathbf a}^{r}_{\mathcal R})$
\STATE \hspace{\algorithmicindent}$+\;\ell_1(\mathbf y,\hat{\mathbf a})$
\STATE $\mathcal L\leftarrow\lambda_a\mathcal L_{\mathrm{act}}+\lambda_s\mathcal L_{\mathrm{score}}+\lambda_{\mathrm{reg}}\mathcal L_{\mathrm{reg}}$
\STATE update HCAT and UGSR by minimizing $\mathcal L$
\STATE \textbf{Inference: return} $\hat{\mathbf a}$ (skip training losses)
\end{algorithmic}
\end{revision}
\end{algorithm}

\insertpagefourfigures
\end{revision}

\subsection{\rev{Uncertainty-guided Span Regeneration}}
\label{sec:selective_regen}
\begin{revision}
HCAT produces a history-conditioned action chunk and per-step uncertainty scores. UGSR uses these outputs to localize and regenerate an unreliable span while retaining the surrounding predictions, as illustrated in Fig.~\ref{fig:method}(b). We evaluate windows of $L$ consecutive future positions and select the eligible window with the highest mean score. Actions within this interval are masked, while the historical prefix and remaining predicted actions stay visible. The refinement decoder then reconstructs the masked actions conditioned on this partially observed tape and the observation context $c_t$. As shown in Fig.~\ref{fig:method}(c), the reconstructed actions are inserted into the selected interval without changing the surrounding predictions. This jointly updates neighboring commands using the executed history and retained future predictions.\fullerpar

For multiple regeneration passes, we recompute uncertainty scores from the updated decoder features after each pass. Subsequent windows are selected from previously unedited positions, and regeneration continues until the specified number of passes is reached or no eligible window remains. Consequently, the final chunk differs from the coarse prediction only within the selected intervals. Algorithm~\ref{alg:cmap} summarizes window selection, masked reconstruction, and interval replacement during policy training and inference.\fullerpar
\end{revision}

\subsection{\rev{Training and Inference}}
\label{sec:training_inference}
\begin{revision}
\nopagebreak[4]
During training, we perform one regeneration pass and supervise the coarse chunk, reconstructed interval, and merged prediction with action reconstruction losses. The uncertainty head receives a detached target computed from the per-step coarse-action error. In Algorithm~\ref{alg:cmap}, $\delta(y,a)$ denotes the action error and $\ell_s(v,r)$ the score loss for prediction $v$ and detached target $r$. On RoboTwin, $\delta(y,a)=\|y-a\|_2^2/d_a$ and $\ell_s(v,r)=e^{-v}r+v$, so $v$ models log expected squared error. On ManiSkill, $\delta(y,a)=\|y-a\|_1/d_a$ and $\ell_s(v,r)=(v-r)^2$, so $v$ models expected absolute error. Both supervise uncertainty from coarse-action errors using the same training procedure. The total objective combines action reconstruction, score supervision, and latent KL regularization $\mathcal L_{\mathrm{reg}}$ adopted from ACT~\citep{zhao2023act}, with $\lambda$ coefficients weighting each loss.\fullerpar

At inference, the policy generates and refines an action chunk from the available observations and executed history. Temporal aggregation combines overlapping predictions for the same execution timestep with exponentially decaying weights~\citep{zhao2023act}. The aggregated command is issued to the robot and appended to the executed history, providing executed-action context for the next policy query and action prediction.\fullerpar
\end{revision}

\setcounter{topnumber}{2}
\insertrevisionrulefigure
\insertcomparisonspage
\insertcomparisoncolumns
\insertcomponenttablepage
\insertsensitivityfigure
\insertlocalregenerationfigure
\insertrealworldrollouts
\insertrealworldtable
\begin{figure}[t]
\cmapfloatat{7}{1}
\centering
\input{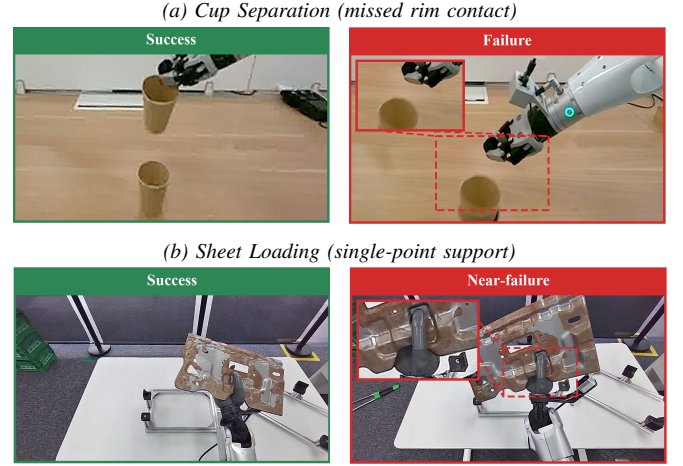}
\setlength{\abovecaptionskip}{2pt}
\caption{\rev{Cup separation on AGIBOT G1 and sheet loading on Kuavo 4 Pro. The left column shows successes. The right column shows a failure and a recovered near-failure. Red callouts mark critical contact regions.}}
\label{fig:realworld_failures}
\end{figure}

\section{Simulation Experiments}
\label{sec:simulation}

\begin{revision}
Our simulations ask how CMAP compares with existing policies and how HCAT and UGSR contribute to task success.\fullerpar
\end{revision}

\subsection{Simulation Benchmarks}
\label{sec:simulation_benchmarks}

\begin{revision}
We evaluate CMAP on six RoboTwin tasks~\citep{mu2024robotwin} and three single-arm ManiSkill tasks~\citep{gu2023maniskill2}, covering diverse manipulation behaviors and interaction requirements. The RoboTwin suite contains \emph{Handover Mic}, \emph{Stack Two Blocks}, \emph{Open Microwave}, \emph{Click Alarm Clock}, \emph{Open Laptop}, and \emph{Place Empty Cup}, covering handover, alignment, contact triggering, articulated-object manipulation, and placement. ManiSkill contains \emph{Push Cube}, \emph{Stack Cube}, and \emph{Pull Cube}. Fig.~\ref{fig:simulation_tasks} illustrates the task scenes and objects in both benchmarks used in this study.\fullerpar
\end{revision}

\subsection{Baselines}
\label{sec:simulation_baselines}

\begin{revision}
\nopagebreak[4]
On RoboTwin, we compare CMAP with task-trained ACT~\citep{zhao2023act}, Diffusion Policy (DP)~\citep{chi2025diffusion}, and Discrete Diffusion Policy (DDP). ACT predicts action chunks with temporal aggregation, DP generates continuous actions through denoising, and our DDP baseline follows MGP's confidence-guided remasking~\citep{mgp} to refine discrete action tokens~\citep{oord2017vqvae} before decoding them into continuous commands. We additionally compare with RDT-1B~\citep{liu2024rdt} and TwinVLA~\citep{im2026twinvla} using the task-level success rates reported in their original papers.\fullerpar

On ManiSkill, ACT uses the same demonstration split as CMAP. OpenVLA~\citep{kim2024openvla} and StaMo~\citep{liu2026stamo} provide additional comparisons on all three single-arm manipulation tasks.\fullerpar
\end{revision}

\subsection{Evaluation Metrics}
\label{sec:simulation_metrics}

\begin{revision}
We report task success rate (\%), the fraction of evaluation episodes that satisfy the benchmark's task-success criterion. On RoboTwin, we evaluate ACT, DP, DDP, and CMAP over 100 episodes per task. The ManiSkill comparison uses the same episode count for ACT and CMAP. Benchmark averages are computed as the unweighted mean of task success rates.\fullerpar
\end{revision}

\subsection{Performance Comparison}

\paragraph{RoboTwin Main Comparison}

\begin{revision}
Across the six tasks in Table~\ref{tab:robotwin_main}, CMAP attains the best or tied-best result on all six tasks and raises the macro-average to 71.7\%, exceeding ACT, DP, and DDP by 21.5, 36.3, and 34.7 percentage points, respectively. Against the strongest task-wise result among ACT, DP, and DDP, its largest gains occur on \emph{Open Microwave} (+27 points), \emph{Open Laptop} (+25), and \emph{Click Alarm Clock} (+13). These tasks require coordinated actions during a brief contact or object-opening phase. HCAT conditions the prediction on preceding commands, while UGSR regenerates a selected interval and retains the surrounding actions. \emph{Stack Two Blocks} requires coordinated grasping, transport, and precise pose alignment. Errors can extend across these stages and exceed a single regeneration span. CMAP achieves 26.0\% success on this task, tying the highest success rate in Table~\ref{tab:robotwin_main}.\fullerpar

DDP achieves 93.0\% success on \emph{Handover Mic}, approaching CMAP, but trails CMAP on the remaining tasks, with particularly large gaps on \emph{Open Microwave} and \emph{Place Empty Cup}. This pattern shows that discrete action revision performs well on handover, while CMAP offers broader gains across manipulation tasks. DDP selects revisions using discrete-token confidence, whereas CMAP uses error-supervised uncertainty to localize and regenerate contiguous spans of continuous actions. Together with the controlled ablations below, these results support localized continuous revision that corrects the relevant commands during contact-sensitive phases while retaining the preceding approach and subsequent continuous action predictions outside the selected interval unchanged.\fullerpar
\end{revision}

\paragraph{ManiSkill Comparison}
\begin{revision}

CMAP averages 53.3\% on ManiSkill, compared with 26.0\% for ACT. The gain over ACT is concentrated in \emph{Pull Cube} (88.0\% versus 34.0\%) and \emph{Push Cube} (54.0\% versus 31.0\%), where the main correction is concentrated around establishing contact and maintaining a largely translational motion. On \emph{Stack Cube}, CMAP achieves 18.0\% success. The task requires coordinated grasping, lifting, transfer, and final pose alignment. The smaller gain is consistent with errors spanning multiple motion stages, beyond the selected regeneration span.\fullerpar
\end{revision}

\subsection{Ablation Studies}

\paragraph{Core Components}
\begin{revision}
Table~\ref{tab:component_buildup} separates the contributions of executed-action history and span selection on three RoboTwin tasks and on ManiSkill. The UGSR column denotes uncertainty-guided span selection. Adding history improves success on both benchmarks, providing a temporal reference for continuing the motion already issued to the robot. Random-span regeneration produces a small additional gain on RoboTwin but reduces success on ManiSkill. Thus, regeneration alone does not consistently improve performance. Uncertainty-guided selection achieves the highest success on both benchmarks. Executed history and uncertainty-guided span selection therefore provide complementary gains.\fullerpar

Fig.~\ref{fig:ablation} compares four action-revision rules on three tasks. Refine-block applies residual corrections within a contiguous block, whereas Refine-uncert corrects individually selected high-uncertainty timesteps. RandSpan and CMAP regenerate masked spans, using random and uncertainty-guided selection, respectively. CMAP performs best on all three tasks, with a macro-average advantage of 15.7 percentage points over the strongest alternative. Opening a microwave or laptop requires maintaining contact as the articulated object moves, whereas pressing an alarm clock requires alignment followed by actuation. The observed advantage supports uncertainty-guided regeneration of neighboring commands while preserving the surrounding predictions, linking local contact adjustments to the preceding approach and subsequent manipulation motions.\fullerpar
\end{revision}

\paragraph{Configuration Sensitivity}
\begin{revision}
\widowpenalty=10000
\brokenpenalty=10000
Fig.~\ref{fig:cmap_sensitivity} evaluates \emph{Open Microwave}, \emph{Open Laptop}, and \emph{Click Alarm Clock}. The default configuration uses $L=8$, $P=1$, and $\rho=0$, with temporal aggregation enabled. Here, $\rho$ denotes the probability of masking historical actions during training. Each sweep varies one setting while keeping the remaining settings fixed. The pass-count sweep varies inference-time regeneration, and the aggregation comparison uses per-step policy replanning.\fullerpar

Keeping the full history ($\rho=0$) achieves 83.7\%, compared with 74.3\% and 67.7\% at mask rates of 0.3 and 0.6. The decline with stronger masking supports preserving executed commands as temporal context. One regeneration pass raises success from 68.3\% to 83.7\%, while two passes achieve 75.7\%. A single pass therefore yields the highest success in this sweep while requiring only one local reconstruction.\fullerpar

Increasing span length from eight to twelve and sixteen steps reduces success from 83.7\% to 68.3\% and 61.3\%, respectively. Shorter spans retain more coarse actions as context and concentrate correction within a smaller region. With per-step replanning, enabling temporal aggregation raises success from 0.0\% to 83.7\%. This execution setting combines overlapping predictions across policy updates, complementing local revision within each chunk with aggregation of overlapping commands across successive policy updates.\fullerpar
\end{revision}

\paragraph{Local Regeneration Analysis}
\begin{revision}
Fig.~\ref{fig:local_regeneration} examines local regeneration on ten held-out demonstration trajectories per task. Span MSE averages squared differences between predicted and expert normalized actions over all timesteps and action dimensions in the selected span. Panel (a) plots the percentage of windows whose span MSE does not exceed each threshold on the logarithmic horizontal axis. The lower curves show After-minus-Before differences in percentage points, with positive values indicating that more windows fall below the same error threshold after regeneration. Within the same selected spans, one regeneration pass reduces mean normalized-action MSE by 21.7\%, 22.1\%, and 39.3\% on \emph{Open Microwave}, \emph{Click Alarm Clock}, and \emph{Open Laptop}, respectively. These reductions show that a single regeneration pass improves the selected action segments while leaving all surrounding predicted actions unchanged by refinement.\fullerpar

To evaluate span selection, Fig.~\ref{fig:local_regeneration}(b) compares Guided, Fixed Span, and Random using the same model checkpoints and span length. Fixed Span uses a centrally placed interval held constant across predictions, while Random averages over all equal-length candidate intervals. For each trajectory, we compute the ratio between post-regeneration and coarse-prediction MSE after averaging over its evaluated windows. Each connected set of markers represents the same held-out trajectory evaluated under the three selection strategies. The reported mean gives equal weight to each trajectory. Guided yields mean ratios below one on all three tasks, whereas both controls yield ratios above one. The same regeneration procedure applied to fixed or random intervals does not match the error reduction from guided selection on these tasks.\fullerpar
\end{revision}

\section{Real-world Experiments}
\label{sec:realworld}
\nopagebreak[4]

\subsection{Real Robot Benchmark}
\begin{revision}
\nopagebreak[4]
We compare CMAP with ACT, DP, and DDP on four real-world tasks across two robot platforms. AGIBOT G1 performs \emph{Cup Separation}, \emph{Bell Press}, and \emph{Dual-item Placement}, while Leju Kuavo 4 Pro performs long-horizon \emph{Sheet Loading}. These tasks examine contact-sensitive manipulation and sustained grasp stability during object transport and placement. Both platforms use RGB observations for visual feedback. Figs.~\ref{fig:realworld_rollouts} and~\ref{fig:realworld_failures} show task executions, final outcomes, and representative contact-related failure and recovery cases.\fullerpar
\end{revision}

\subsection{Deployment Protocol}
\begin{revision}
\nopagebreak[4]
All four policies use the same task-specific teleoperated datasets, with 50 demonstrations for each AGIBOT task and 150 for \emph{Sheet Loading}. Observations, action spaces, and evaluation settings are matched across policies. Each policy is evaluated in ten trials per task, totaling 40 rollouts.\fullerpar

\emph{Cup Separation}, \emph{Bell Press}, and \emph{Sheet Loading} are scored by complete task success. For \emph{Dual-item Placement}, chess-grasp and bottle-grasp success are recorded separately within the same ten trials. Table~\ref{tab:realworld_main} reports five success counts out of ten and their equally weighted mean success rate.\fullerpar
\end{revision}

\subsection{Performance Comparison}
\begin{revision}

On AGIBOT G1, CMAP achieves the highest success across all four evaluation measures, averaging 85.0\%, compared with 57.5\% for both ACT and DP. The bottle-grasp stage shows its largest advantage over DDP, with CMAP succeeding in all ten trials and DDP in five. These tasks require coordinated transitions from approach to contact and subsequent manipulation, particularly the alignment of gripper closure or pressing with the approach motion. CMAP uses executed commands as a temporal reference for new predictions and jointly revises neighboring commands while retaining the surrounding actions. These gains are consistent with maintaining action continuity during contact-sensitive manipulation.\fullerpar

On Leju Kuavo 4 Pro, CMAP completes \emph{Sheet Loading} in 4/10 trials, compared with 2/10 for the strongest baselines. Unlike brief grasping or pressing interactions, sheet loading requires sustained grasp stability through lifting, transport, and fixture alignment. Establishing the initial grasp is only the first step, as subsequent motion must coordinate the workpiece pose with its support conditions. Local regeneration provides focused corrections within the predicted action chunk, while receding-horizon updates allow subsequent predictions to respond to changes in object pose. The recovery example in Fig.~\ref{fig:realworld_failures} further illustrates adjustment during execution, while the remaining failures highlight the challenge of maintaining support and alignment across motion phases. Across the five measures in Table~\ref{tab:realworld_main}, CMAP achieves a mean success rate of 76.0\%, with gains spanning local contact interactions on AGIBOT and complete sheet-loading execution on Kuavo.\fullerpar
\end{revision}

\paragraph{Comparison with DDP}
\begin{revision}
The DDP comparison evaluates two ways of selecting actions for revision. DDP uses categorical-token confidence to select tokens for remasking, whereas CMAP aggregates error-supervised uncertainty over continuous action positions to select a contiguous span. The real-world advantage over DDP complements the simulation ablations, which separately examine executed history, span regeneration, and score-guided selection. Together, these results support combining history-conditioned prediction with temporally coherent local revision for contact-sensitive manipulation on both evaluated robot platforms.\fullerpar
\end{revision}

\paragraph{Failure and Recovery Analysis}
\begin{revision}
Fig.~\ref{fig:realworld_failures} contrasts successful execution with a cup-separation failure and a recovered sheet-loading near-failure. The failure occurs when the gripper closes without securing the cup rim, showing the importance of coordinating gripper alignment and closure before lifting.\fullerpar
\end{revision}

\begin{revision}
In \emph{Sheet Loading}, the near-failure frame shows the workpiece supported at a single point, in contrast to the more stable support in the successful example. The rollout subsequently recovers through replanning. This case illustrates a different correction requirement from cup separation. The robot must maintain support while adjusting the workpiece pose, so a useful correction must coordinate the ongoing motion with the contact that remains available. Together, the examples highlight both timely contact establishment and sustained support during manipulation. They motivate incorporating contact information into span selection and adapting the revision interval to the correction duration of the task.\fullerpar
\end{revision}
\section{Conclusion}
\label{sec:conclusion}

\begin{revision}
We presented CMAP for history-conditioned local regeneration of continuous action chunks. HCAT incorporates executed commands, while UGSR jointly regenerates an uncertainty-guided contiguous span and preserves all remaining actions. CMAP achieves mean success rates of 71.7\% on RoboTwin, 53.3\% on ManiSkill, and 76.0\% across the five real-world evaluation measures. Ablations demonstrate the complementary roles of executed history and uncertainty-guided span regeneration, while held-out analysis shows reduced errors within regenerated spans. Future work may incorporate contact information and adaptive spans to support local action regeneration during long-horizon robotic manipulation.\fullerpar
\end{revision}

\bibliographystyle{IEEEtran}
\bibliography{ref}

\end{document}